\documentclass[letterpaper]{article} 
\usepackage{aaai23}  
\usepackage{times}  
\usepackage{helvet}  
\usepackage{courier}  
\usepackage[hyphens]{url}  
\usepackage{graphicx} 
\usepackage{natbib}  
\usepackage{caption} 
\usepackage{algorithm}
\usepackage{algorithmic}

\usepackage{newfloat}
\usepackage{listings}
\DeclareCaptionStyle{ruled}{labelfont=normalfont,labelsep=colon,strut=off} 
\floatstyle{ruled}
\newfloat{listing}{tb}{lst}{}
\floatname{listing}{Listing}
\usepackage[inline]{enumitem}
\usepackage{mathtools}
\usepackage{amsmath}
\usepackage{booktabs}
\usepackage{amssymb}
\usepackage{multirow}

\title{Improving the Cross-Lingual Generalisation in Visual Question Answering}
\author {
    Farhad Nooralahzadeh,
    Rico Sennrich
}
\affiliations {
    Department of Computational Linguistics, University of Zurich\\
    fahrad.nooralahzadeh@uzh.ch,
    sennrich@cl.uzh.ch
}

\usepackage{bibentry}

\begin{document}

\maketitle

\begin{abstract}
While several benefits were realized for multilingual vision-language pretrained models, recent benchmarks across various tasks and languages showed poor cross-lingual generalisation when multilingually pre-trained vision-language models are applied to non-English data, with a large gap between (supervised) English performance and (zero-shot) cross-lingual transfer.
In this work, we explore the poor performance of these models on a zero-shot cross-lingual visual question answering (VQA) task, where models are fine-tuned on English visual-question data and evaluated on 7 typologically diverse languages.
We improve cross-lingual transfer with three strategies:
\begin{enumerate*}[label=(\arabic*), itemjoin={{, }}, itemjoin*={{, and }}]
    \item we introduce a linguistic prior objective to augment the cross-entropy loss with a similarity-based loss to guide the model during training
    \item we learn a task-specific subnetwork that improves cross-lingual generalisation and reduces variance without model modification

    \item we augment training examples using synthetic code-mixing to promote alignment of embeddings between source and target languages. 
\end{enumerate*} 
Our experiments on xGQA using the pretrained multilingual multimodal transformers UC2 and M3P demonstrate the consistent effectiveness of the proposed fine-tuning strategy for 7 languages, outperforming existing transfer methods with sparse models.

\end{abstract}
\begin{figure*}[ht]
\begin{minipage}[b]{0.20\linewidth}
\centering
\includegraphics[width=1\textwidth]{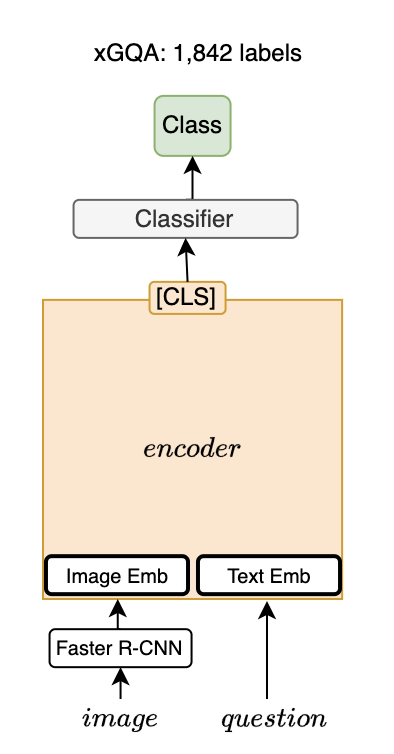}
\caption{A standard setup \citep{DBLP:journals/corr/abs-2201-11732, pfeiffer-etal-2022-xgqa} to perform VQA task using UC2 or M3P.}
\label{fig:design-0}
\end{minipage}
\hspace{0.3cm}
\begin{minipage}[b]{0.80\linewidth}
\centering
\includegraphics[width=1\textwidth]{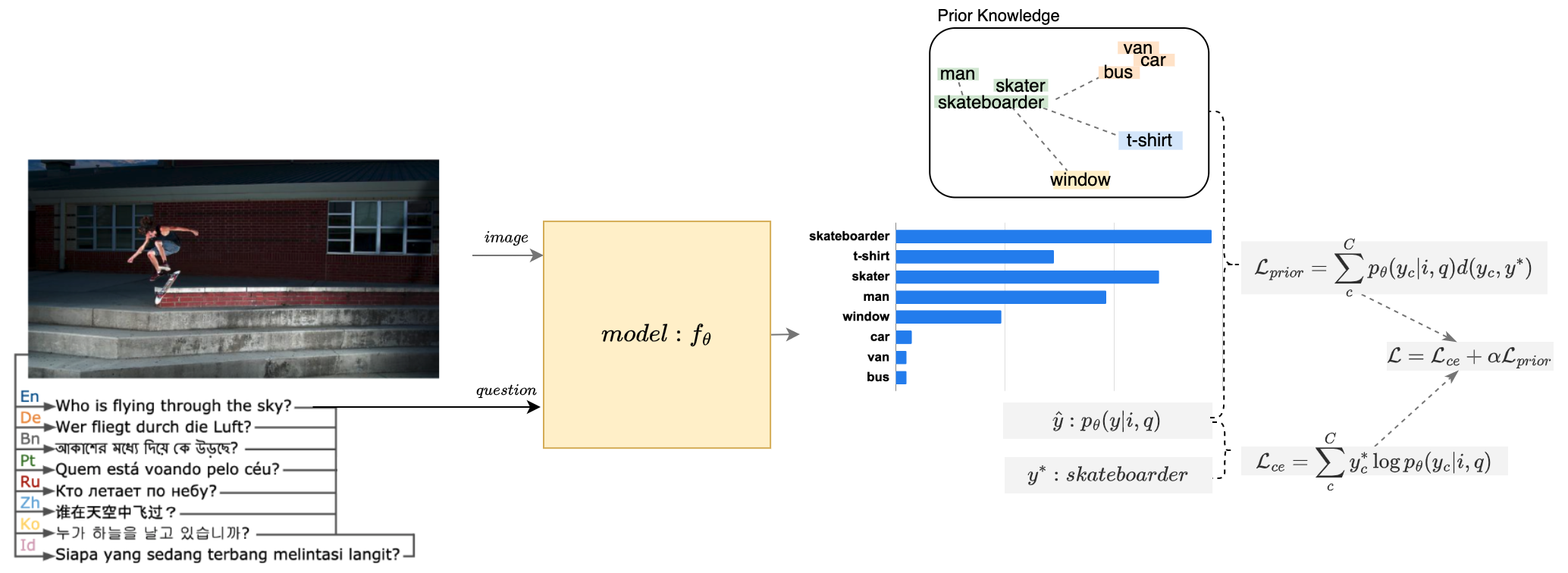}
\caption{Illustration of how we augment the cross-entropy loss with a similarity-based loss using a linguistic prior knowledge in VQA task. The model receives a question-image pair where the question is \textit{Who is flying through the sky?} and the ground truth label is \textit{skateboarder}. Example is taken from xGQA \citep{pfeiffer-etal-2022-xgqa} dataset. }
\label{fig:design-1}
\end{minipage}
\end{figure*}
\section{Introduction}
 Multimodal pretraining has established state-of-the-art performance for many multimedia tasks such as image-text retrieval, visual question, and answering, video localization, speech recognition, etc. Pretraining models outperforms traditional methods by providing stronger representation of different modalities learned in an unsupervised training fashion \citep[e.g.][]{pmlr-v139-radford21a,schneider19_interspeech,Sun2019VideoBERTAJ}. However, progress in this area has been limited mostly to the English language, whereas the main multimodal datasets consist only of English data. In order to generalize this achievement to non-English languages, recent works \citep[e.g.][]{zhou2021uc, 9577347,liu-etal-2021-visually, bapna2022mslam} attempt to learn universal representations to map objects that occurred in different modalities or texts expressed in various languages into shared semantic space. 

IGLUE \citep{DBLP:journals/corr/abs-2201-11732}, a recent benchmark spanning various tasks and languages has shown that performance degrades significantly when existing multilingual vision-language models are applied  to  non-English data, and there is a large gap between supervised performance and (zero-shot) cross-lingual transfer.
This gap is most noticeable for resource-poor languages and languages that are distinct from English, attributed mostly to misalignment of text embeddings between the source and target languages \citep{liu-etal-2021-visually,pfeiffer-etal-2022-xgqa}. 

In this work, we address a number of deficiencies in how these multilingual vision-language models are trained and evaluated on xGQA \citep{pfeiffer-etal-2022-xgqa}, a multilingual evaluation benchmark for the visual question answering task, where the source English dataset is extended to 7 typologically diverse languages.
Specifically, we address the following issues:
\begin{enumerate*}[label=(\roman*), itemjoin={{, }}, itemjoin*={{, and }}]
    \item The standard cross-entropy loss function fails to assess properly the different incorrect model outputs and results in treating equally all incorrect predictions during training
    \item The label space is highly derived from the source language (i.e.\ English), resulting in language bias in the training material and hurting generalization to other languages
    \item The non-restricted fine-tuning of multilingual vision-language models likely neglect the task-specific and language-neutral components, resulting in over-fitting on the source language and poor cross-lingual generalisation.
\end{enumerate*}
Our contributions are as follows: 
\begin{enumerate}[leftmargin=*]
    \item We design an effective fine-tuning strategy by incorporating the linguistic prior, task-specific sparse sub-network, and synthetic code-mixing augmentation  to address the low performance of pretrained multilingual vision-language models on cross-lingual VQA task. Our strategy does not introduce extra trainable parameters or layers, and even reduces the number of model parameters. Code and data to reproduce our findings are publicly available.\footnote{
\url{https://github.com/nooralahzadeh/CLG-VQA}}
    
    \item We evaluate the proposed strategy on cross-lingual zero-shot learning, across a total of 7 languages and observe consistent improvements over strong multilingual multimodal transformers including UC2 \cite{zhou2021uc} and M3P  \cite{9577347}, achieving a substantial $+13.12$\% and $+12.63$\% gain in average accuracy over all languages in xGQA against UC2 and M3P baselines, respectively.
    
    \item We perform an error analysis highlighting a substantial number of confusions between semantically related labels in xGQA, including synonyms, hypernyms, and hyponyms. We propose a metric that treats all synonyms of the ground truth label as correct.

\end{enumerate}
\section{Background} \label{sec:background}
Having a pair of an image and a question, the task in visual question answering (VQA) is to provide an answer considering both modalities. This process has been treated as a classification task in most VQA benchmark datasets, where the underlying model should select one or multiple answers from a set of predefined labels. Recently, \citet{pfeiffer-etal-2022-xgqa} introduce a typologically diverse multilingual and multimodal benchmark for VQA task by extending the monolingual English-only GQA \citep{Hudson2019GQAAN} dataset. They utilize 12,578 questions and 398 images from the test and development set of GQA, where the questions are manually translated into 7 different languages, covering 5 different scripts: Bengali (Bn), German (De), Indonesian (Id), Korean (Ko), Portuguese (Pt), Russian (Ru) and simplified Chinese (Zh).  
The xGQA benchmark also consists of new fixed data splits to guide cross-lingual few-shot learning experiments, where only a small number of examples in the target language are available.
This dataset has been used in recent studies on cross-lingual transfer
learning of vision-language models \citep[e.g.][]{Liu2022DelvingDI,Zeng2022CrossViewLM} and includes several types of structured questions about an image. 
In this work, we base our approach on  two state-of-the-art pretrained multilingual vision-language architectures, namely UC2 \cite{zhou2021uc} and M3P \cite{9577347}.
These two transformer-based multimodal models accept the concatenation of image region features extracted with an object detector (i.e. Faster R-CNN  \citep{DBLP:conf/nips/RenHGS15}) and a sequence of BPE tokens \citep{sennrich-etal-2016-neural} representing the question using Sentence Piece model \citep{kudo-richardson-2018-sentencepiece} as an input. This input is then processed by a BERT-like encoder \citep{devlin-etal-2019-bert} to obtain multimodal, contextualised representations. They are initialized from XLM-R \citep{DBLP:journals/corr/abs-1911-02116} and mainly differ in 
their pretraining strategy.

As Figure~\ref{fig:design-0} depicts, the standard setup \citep{pfeiffer-etal-2022-xgqa,DBLP:journals/corr/abs-2201-11732} to perform cross-lingual VQA task is to fine-tune the pretrained multilingual image-text model in the source language (i.e, English). Then, the representation of the [CLS] token as a multimodal and contextualized representation is fed into a non-linear two-layer feed-forward classifier head to predict an answer for a given image-question pair. For \textit{zero-shot cross-lingual} evaluation, the fine-tuned model is evaluated on the multilingual test data, whereas in \textit{few-shot cross-lingual} scenario, the fine-tuned model is additionally trained on image-question examples available in the target language. 
\section{Fine-Tuning Strategies}
Multilingual vision-language pretrained models often suffer from poor cross-lingual generalisation compared with their corresponding monolingual baseline, achieving much better performance in the source language than in target languages unseen during fine-tuning.
In this work, we aim to address the poor performance of these models on the xGQA benchmark, where models are fine-tuned on English data and evaluated on 7 typologically diverse languages.
In particular, we investigate the impact of three fine-tuning strategies:
\begin{enumerate*}[label=(\roman*), itemjoin={{, }}, itemjoin*={{, and }}]
    \item Incorporating linguistic prior
    \item Task-specific sparse fine-turning
    \item Multilingual Code-Switching (i.e. Code-Mixing) data augmentation.
\end{enumerate*}
In this section, we describe these three strategies in detail.
\subsubsection{Incorporating Linguistic Prior} \label{sec:prior}
We realize a number of deficiencies in how multilingual vision-language models are trained and evaluated cross-lingually in the VQA task.
\begin{enumerate*}[label=(\roman*), itemjoin={{, }}, itemjoin*={{, and }}]
    \item The loss function fails to assess properly the different
incorrect model outputs and results in treating equally all
incorrect predictions during training
\item Training examples are only annotated with one label, where intuitively multiple labels are often plausible (e.g.\ lady vs.\ woman; couch vs.\ sofa)
\item The label space highly depends on the source language
(i.e.\ English) and hurts generalization to other languages. For instance, there are singular and plural labels such as 
\emph{car/cars}, \emph{woman/women} and \emph{laptop/laptops} while in some target languages such as Chinese, most nouns are not marked for grammatical numbers. 
\end{enumerate*} 
In this section, we aim to address these issues.

Given an image $i$ and a question $q$, the respective model $f_{\theta}$ for VQA task provides a probability distribution $\hat{y}=p_{\theta}(y|i,q)$  over a set of predefined answers.
Commonly, VQA models are trained using the cross-entropy loss, in which parameters of the underlying model $\theta$ are optimized using the following objective function: \[\mathcal{L}_{ce}= \sum_{c=1}^{C} y^{*}_{c} \log p_{\theta}(y_{c}|i,q)\] where $C$ is the number of classes in the answer set, and $y^{*}$ is the one-hot vector that represents the ground truth answer. 
The objective loss function encourages the model to give a large probability mass to a correct class. It compares the predicted and ground-truth label and takes a once-for-all matching strategy, consequently evaluating all predictions as either correct or incorrect and ignoring the similarity between the correct and less incorrect predictions.
As an example in the question-image pair shown in Figure~\ref{fig:design-1}, if the model receives a question as \textit{Who is flying through the sky?} and the ground truth label is  \textit{skateboarder}, the underlying loss function will penalize the wrong predicted labels such as  \textit{skater}, \textit{man}, \textit{t-shirt} or  \textit{car} equally.
We argue that the incorrect training predictions may be quite diverse and letting the model be aware of which incorrect predictions are more incorrect or less incorrect than others may more effectively guide the model during the training.
Therefore, in our example, similar labels such as  \textit{skater} and  \textit{man} should be penalized much less than dissimilar words like  \textit{t-shirt} or  \textit{car}. 

In order to alleviate the issue, we add a linguistic prior objective to augment the cross-entropy loss with a similarity-based loss. The loss can be conceived as a form of risk minimization, where the risk function is the distance $d$ between a ground truth label $y^{*}$ and the predicted label $y_{c}$. In other words, the objective function should give a small loss if the predicted and ground truth label are similar, and penalize dissimilar answers:

\begin{equation*}
  \begin{split}
\mathcal{L}_{prior} &= \sum_{c} p_{\theta}(y_{c}|i,q) d(y_{c},y^{*})\\
\mathcal{L}&= \mathcal{L}_{ce}+ \alpha \mathcal{L}_{prior}
\end{split}
\end{equation*}

The risk $d$ is weighted by the probability distribution over all target labels $p_{\theta}(y|i,q)$, provided by the classification layer.
We formalize the distance score $d(y_c,y^{*})$ between the ground truth label and others in the label space by using two sources of linguistic knowledge:
\paragraph{WordNet} (prior$_{wn}$):
We extract the explicit relations among the labels using the synset structure of the English lexical database (i.e. WordNet \citep{wordnet}). To be more precise, we derive the synonymy, hyponymy, and hypernymy relations and formulate the distance as:

\[
d(y_c,y^*)=
\begin{cases}
    0  & \text{if } y_c  \text{ and } y^* \text{ are } synonyms   \\
    d_1              &
    \text{if } y_c \text{ is } hyponym  \text{ of } y^*\\
    d_2 & \text{if } y_c \text{ is } hypernym  \text{ of } y^* \\
    1 & \text{otherwise}
\end{cases}
\]

, where { $0<d_1, d_2<1$}.
\paragraph{Word Embeddings} (prior$_{em}$): A distance $d$ is extracted from implicit semantic proximity within pretrained word embeddings. We  calculate the embeddings cosine distance as the distance of $y^*$ and all other labels as: 
\[d(y_c,y^*)=Cosine Distance(emb_{y^*}, emb_{y_c})\]
\subsubsection{Task-specific Sparse Fine-tuning  \textsc{(SfT)}}
The success of multilingual pretrained models in cross-lingual genralisation is often attributed to task-specific and language-neutral components, which capture commonalities among languages
\citep{libovicky-etal-2020-language, Foroutan2022DiscoveringLS}.

To this end, we are inspired by
previous works \cite{frankle2018the, chen2020lottery,ansell-etal-2022-composable} that claim there exists a sparse, separated trainable subnetwork (i.e. a winning ticket) capable to match or even outperform the original neural network.
Similarly, we design a task-specific sparse fine-tuning strategy, here dubbed \textsc{SfT}, consisting of two steps:
\begin{algorithm}[tb]
\caption{Iterative Magnitude Pruning (IMP) with rewinding step \cite{DBLP:journals/corr/HanMD15}.}
\label{alg:imp}
\textbf{Input}: Model $f(.; \theta)$ initialized with pretrained parameters $\theta^0$.\\
\textbf{Parameter}: $p$\% : a pruning rate \\
\textbf{Output}: M
\begin{algorithmic}[1] 
\STATE Set the initial pruning mask to $M = 1^{|\theta|}$.
\WHILE{not done}
\STATE Train $f(.; M \odot \theta^0)$ to step $t$: $f(.; M \odot \theta^t)$.
\STATE Prune $p$\% of remaining weights of $M \odot \theta^t$ and update $M$ accordingly.
\ENDWHILE
\STATE Return $f(.; M \odot \theta^0)$.
\end{algorithmic}
\end{algorithm}

\paragraph{$Step_0$:} Considering the VQA model $f(.;\theta)$ initialized with pretrained weights $\theta^0$, we obtain a subnetwork $f(.;M\odot \theta)$ where $M \in \{0,1\}^{|\theta|}$ represents a binary mask and $\odot$ is element-wise multiplication. More specifically as it is shown in Algorithm~\ref{alg:imp}, we utilize   \textit{Iterative Magnitude Pruning} (IMP) \cite{DBLP:journals/corr/HanMD15} to discover the pruning mask $M$, during the fine-tuning of the VQA model in English-only data. After each epoch, we prune a certain amount (e.g. $p$\%) of the original parameters. Then, we continue the fine-tuning by resetting the remaining parameters to their original value on the pretrained initialization $\theta^0$.\\
\paragraph{$Step_1$:} Having the pruning mask $M$, the model parameters are initialized with their original values $\theta ^0$ and are fine-tuned again. However, in this step, only the unmasked parameters are trained while the masked ones are kept frozen. It should be mentioned that following previous works \cite{ zhou_2019_dlt,chen2020lottery} the masked parameters are set to zero.
\subsubsection{Code-Mixing \textsc{(CdM)}} 
\begin{figure}[t]
    \includegraphics[width=.5\textwidth]{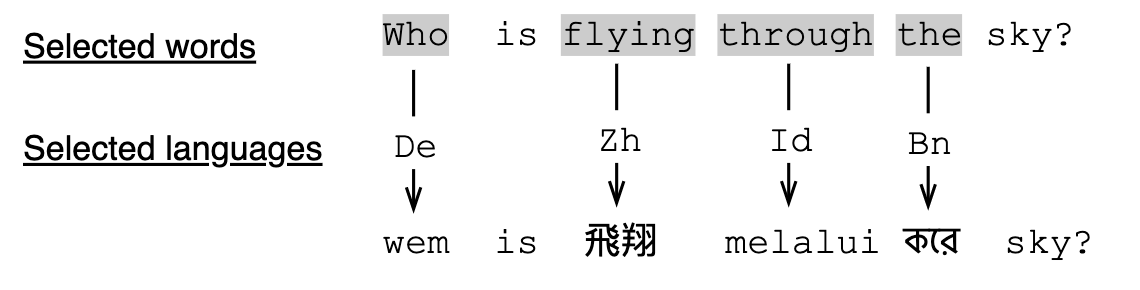}
    \caption{The code-mixed question, where a set of 4 words is randomly selected in order to be replaced by their translation using  bilingual dictionaries of MUSE \citep{DBLP:conf/iclr/LampleCRDJ18} into 4 randomly selected target languages in xGQA.}
    \label{fig:code-mixed}
        \vspace{-0.5cm}

\end{figure}

While each fine-tuning step only involves questions from the English language, the VQA task is unable to benefit properly from cross-lingual alignment information that exists in multilingual vision-language models. To make full use of this cross-lingual alignment information and better fine-tuning, we construct code-mixed data in target languages. To generate the code-mixed questions, we follow the mechanism of multilingual code-switching data augmentation (CoSDA) proposed by \citet{Qin2020CoSDAMLMC}. First, a set of words is randomly chosen in each question. Second, for each selected word, we randomly specify a target language to translate. Third, we replace the word with its translation in the selected language. If the word has multiple translations in the target language, then one of them is randomly selected for replacement.
To increase the data diversity during the training, \citet{Qin2020CoSDAMLMC} proposes to reset the replacement after each epoch and to replace different words at different epochs.\footnote{For further details regarding CoSDA we refer the reader to the original work.} Figure \ref{fig:code-mixed} shows the result of applying the code-mixing procedure to our example.

\section{Experiments}
To evaluate our proposed strategies, as explained in Section \ref{sec:background}, we benchmark two state-of-the-art multilingual vision-language transformers, namely UC2 and M3P, as the base models. We study the impact of each strategy by fine-tuning the model on the monolingual English GQA dataset\footnote{We consider \textit{balanced} subset of GQA as recommended in IGLUE benchmark \cite{DBLP:journals/corr/abs-2201-11732} }, then evaluating the cross-lingual transfer on the multilingual extension of GQA, known as xGQA. We adopt the codebase of IGLUE benchmark\footnote{\url{https://iglue-benchmark.github.io/}} to implement
our proposed approach and we keep the value of the models and training hyper-parameters equal to the ones that are reported by \citet{DBLP:journals/corr/abs-2201-11732}.
The results are reported for each experiment by averaging the performance over five different runs.
\subsection{Model Configurations and Notation}
On both UC2 and M3P models, we experiment with three different setups:
\begin{table*}[tb]
      \resizebox{0.99\textwidth}{!}{
    \centering
    \scriptsize
    \begin{tabular}{c l c ccccccc c}
    \toprule
        & \textbf{Model} & En & Bn & De & Id & Ko & Pt & Ru & Zh & \textbf{Avg}\\
        \midrule
 \multicolumn{11}{c}{\emph{Fine-tune model on English training set (Zero-Shot)}} \\
\midrule
       UC2 & Our Baseline & 
        54.92 &
        19.99 &
        42.00 &
        28.44 &
        22.40 &
        30.92 &
        28.55 &
        31.19 &
        29.07\\
         & Baseline \citep{DBLP:journals/corr/abs-2201-11732} &  
        55.19  &
        19.98  &
        42.85  &
        28.67 &
        21.36  &
        30.41  &
        30.99  & 
        31.15   &
        29.35
       \\
       & \citet{Liu2022DelvingDI} & 
        58.57$\scriptscriptstyle\pm 0.2$ &
        26.23 $\scriptscriptstyle\pm 1.5$ &
        49.51 $\scriptscriptstyle\pm 1.1$ &
       38.92 $\scriptscriptstyle\pm 1.3$ &
       36.48 $\scriptscriptstyle\pm 1.3$ &
        39.76 $\scriptscriptstyle\pm 0.6$ &
        41.72 $\scriptscriptstyle\pm 0.3$ &
        \bf{46.52} $\scriptscriptstyle\pm 0.9$ &
        39.87\\
  \cmidrule(lr){2-2} \cmidrule(lr){3-3} \cmidrule(lr){4-11} 
        & With prior$_{wn}$  &
        55.77$\scriptscriptstyle\pm 0.02$ &
        23.66 $\scriptscriptstyle\pm 0.76$ &
        47.93 $\scriptscriptstyle\pm 0.19$ &
       35.67 $\scriptscriptstyle\pm 1.43$ &
       34.57 $\scriptscriptstyle\pm 1.81$ &
        37.46 $\scriptscriptstyle\pm 1.35$ &
        40.08$\scriptscriptstyle\pm 0.54$ &
        40.08$\scriptscriptstyle\pm 4.31$ &
        37.06\\
     & With prior$_{em}$ & 
        56.09 $\scriptscriptstyle\pm 0.14$ &
        23.97$\scriptscriptstyle\pm 2.56$ &
        48.13 $\scriptscriptstyle\pm 0.78$ &
       36.87$\scriptscriptstyle\pm 1.90$ &
       34.14 $\scriptscriptstyle\pm 3.56$ &
        38.18 $\scriptscriptstyle\pm 2.55$ &
        41.07 $\scriptscriptstyle\pm 0.86$ &
        41.76$\scriptscriptstyle\pm 1.89$ &
        37.73\\
\cmidrule(lr){2-2} \cmidrule(lr){3-3} \cmidrule(lr){4-11} 
        &  With prior$_{em}$+ \textsc{SfT} & 
        56.56$\scriptscriptstyle\pm 0.10$ &
        23.53 $\scriptscriptstyle\pm 1.97 $ &
        49.54 $\scriptscriptstyle\pm 0.27$ &
       36.79 $\scriptscriptstyle\pm 0.46$ &
       34.56 $\scriptscriptstyle\pm 0.49$ &
        38.95 $\scriptscriptstyle\pm 0.19$ &
        41.18 $\scriptscriptstyle\pm 0.23$ &
        43.40 $\scriptscriptstyle\pm 0.21$ &
        38.28\\
         & With prior$_{em}$ + \textsc{CdM} &  							 
        54.37$\scriptscriptstyle\pm 0.01 $ &
        27.38 $\scriptscriptstyle\pm 0.02$ &
        46.66 $\scriptscriptstyle\pm 1.70$ &
        20.88 $\scriptscriptstyle\pm 2.33 $ &
        36.32$\scriptscriptstyle\pm 1.11$ &
        40.81 $\scriptscriptstyle\pm 2.06$ &
        43.48 $\scriptscriptstyle\pm 0.18$ &
        30.62 $\scriptscriptstyle\pm 1.46$ &
        35.16\\
         \cmidrule(lr){2-2} \cmidrule(lr){3-3} \cmidrule(lr){4-11} 
        & With prior$_{em}$ + \textsc{SfT} + \textsc{CdM} &
               55.21$\scriptscriptstyle\pm 0.08$ &
        \bf{30.96} $\scriptscriptstyle\pm 1.33$ &
        \bf{50.30} $\scriptscriptstyle\pm 0.22$ &
       \bf{41.68} $\scriptscriptstyle\pm 0.74$ &
       \bf{39.57} $\scriptscriptstyle\pm 0.65$ &
        \bf{43.43} $\scriptscriptstyle\pm 0.60$ &
        \bf{44.58} $\scriptscriptstyle\pm 0.92$ &
        {44.80} $\scriptscriptstyle\pm 0.78$ &
        \bf{42.19}\\
        \midrule
          \midrule
       M3P & Our Baseline & 
                54.02 &
        17.24 &
        32.40 &
       23.77  &
       25.57 &
        32.91 &
        32.32  &
        27.39  &
        27.37\\
      & Baseline \citep{DBLP:journals/corr/abs-2201-11732} &  
      53.75   &
      18.64   &
      33.42    &
      32.48   &
      25.11    &
      31.40     &
      27.50    &
      28.65    &
      28.17
       \\
       &\citet{Liu2022DelvingDI} & 
        46.70$\scriptscriptstyle\pm 0.7$ &
        29.75 $\scriptscriptstyle\pm 1.4$ &
        39.52 $\scriptscriptstyle\pm 1.3$ &
       {36.73} $\scriptscriptstyle\pm 1.6$ &
       35.67 $\scriptscriptstyle\pm 1.1$ &
        37.59 $\scriptscriptstyle\pm 0.8$ &
        37.93 $\scriptscriptstyle\pm 0.9$ &
        36.15 $\scriptscriptstyle\pm 0.9$ &
        36.19\\
  \cmidrule(lr){2-2} \cmidrule(lr){3-3} \cmidrule(lr){4-11} 
        & With prior$_{wn}$ &
        55.91$\scriptscriptstyle\pm 0.20$ &
        22.38 $\scriptscriptstyle\pm 0.38$ &
        39.48 $\scriptscriptstyle\pm 1.73$ &
       29.31 $\scriptscriptstyle\pm 2.27$ &
       35.15$\scriptscriptstyle\pm 0.86$ &
        39.00 $\scriptscriptstyle\pm 0.17$ &
        38.92 $\scriptscriptstyle\pm 0.31$ &
        35.74 $\scriptscriptstyle\pm 0.79$ &
        34.28\\
     & With prior$_{em}$ & 
        56.33$\scriptscriptstyle\pm 0.09$ &
        22.93 $\scriptscriptstyle\pm 3.19$ &
        40.10 $\scriptscriptstyle\pm 0.54 $ &
       30.63 $\scriptscriptstyle\pm 0.05 $ &
       35.35 $\scriptscriptstyle\pm 2.14$ &
        38.85 $\scriptscriptstyle\pm 1.09$ &
        39.95 $\scriptscriptstyle\pm 0.03$ &
        36.97 $\scriptscriptstyle\pm 0.07$ &
        34.97\\
\cmidrule(lr){2-2} \cmidrule(lr){3-3} \cmidrule(lr){4-11} 
     &  With prior$_{em}$ + \textsc{SfT} &    
              56.18 $\scriptscriptstyle\pm 0.00$ &
        22.07 $\scriptscriptstyle\pm 0.46$ &
       40.29 $\scriptscriptstyle\pm 0.17$ &
       27.04 $\scriptscriptstyle\pm 0.11$ &
       34.62 $\scriptscriptstyle\pm 0.06$ &
        38.39 $\scriptscriptstyle\pm 0.24$ &
        39.44 $\scriptscriptstyle\pm 0.00$ &
        36.32 $\scriptscriptstyle\pm 0.49$ &
        34.02\\
         &  With prior$_{em}$ + \textsc{CdM} &    
        54.35$\scriptscriptstyle\pm 0.64$ &
        28.71 $\scriptscriptstyle\pm 1.73$ &
        43.57 $\scriptscriptstyle\pm 0.33$ &
       \bf{38.89} $\scriptscriptstyle\pm 2.07$ &
       38.06 $\scriptscriptstyle\pm 0.44$ &
        41.93 $\scriptscriptstyle\pm 0.59$ &
        41.64 $\scriptscriptstyle\pm 1.11$ &
        38.80 $\scriptscriptstyle\pm 1.22$ &
        38.80\\ \cmidrule(lr){2-2} \cmidrule(lr){3-3} \cmidrule(lr){4-11} 
        &  With prior$_{em}$ + \textsc{SfT} + \textsc{CdM}    & 
                55.58$\scriptscriptstyle\pm 0.12$ &
        \bf{31.53} $\scriptscriptstyle\pm 1.47$ &
        \bf{46.19} $\scriptscriptstyle\pm 0.54 $ &
       34.60 $\scriptscriptstyle\pm 0.49$ &
       \bf{40.21} $\scriptscriptstyle\pm 0.91$ &
        \bf{42.87} $\scriptscriptstyle\pm 0.58$ &
        \bf{42.32} $\scriptscriptstyle\pm 1.19$ &
        \bf{42.25} $\scriptscriptstyle\pm 0.60$ &
        \bf{40.00}\\
        \midrule
     \multicolumn{11}{c}{\emph{Translate everything to English and use the English-only model (Translate-Test)
}} \\
        \midrule
UC2 & \citep{DBLP:journals/corr/abs-2201-11732} & 55.19 &	49.31&	52.61&	50.34&	48.62&	52.17&	49.95&	48.32&	50.19 \\
M3P &\citep{DBLP:journals/corr/abs-2201-11732} & 53.75	& 47.79&	51.01&	49.35&	47.64&	51.21&	47.76&	47.04&	48.83   
\\
    \bottomrule
    \end{tabular}
    }
    \caption{Accuracy results on the xGQA test set for Zero-Shot transfer. Columns indicate the target languages.  We also report the average (Avg.) accuracy across languages excluding English. For our baseline, we fine-tuned the model on the English balanced subset of GQA and evaluated it on the test set of xGQA. In \textit{With prior$_{xx}$}, the original cross-entropy loss is augmented with a similarity-based loss, either using WordNet (i.e. \textit{With prior$_{wn}$}) or Word Embeddings  (i.e. \textit{With prior$_{em}$)}. In \textit{With prior$_{em}$}+\textsc{SfT}, we apply task-specific sparse fine-tuning strategy along with Word Embeddings based loss. In \textit{With prior$_{em}$}+\textsc{SfT}+\textsc{CdM} is our final design where we employ the code-mixing augmentation on top of the previous strategy. For comparison, we report results from  \citet{DBLP:journals/corr/abs-2201-11732} and \citet{Liu2022DelvingDI}. The performance of all proposed strategies is averaged from five runs with different random seeds. }
    \label{tab:main}
     \vspace{-0.5cm}

\end{table*}
\paragraph{With prior$_{xx}$:}
The fine-tuning process is performed using a similarity-based loss together with cross-entropy loss.
The prior knowledge based distances $d$ are computed as follows: \begin{enumerate*}[label=(\roman*), itemjoin={{, }}, itemjoin*={{, and }}]
\item \emph{prior$_{wn}$}: The WordNet base distance is computed by using the NLTK library \citep{journals/corr/cs-CL-0205028}
\item \emph{prior$_{em}$}: To compute the cosine distance among 1842 labels in xGQA, we use the spaCy\footnote{\url{https://spacy.io/}} toolkit, where an embedding $emb_y \in \mathbb{R}^{300}$ of each label is derived from GloVe \citep{pennington-etal-2014-glove} pretrained word embeddings.
\end{enumerate*} 

To mitigate the negative influence of non-probable classes on the similarity-based loss, we consider the $k$ most probable answers according to their probability $p_{\theta}(y_c|i,q)$ in both setups.
We set hyper-parameters $d_1=0.8$, $d_2=0.8$, $k=10$ and $\alpha =10$ based on validation set performance.\footnote{We performed a grid search using different values for these hyper-parameters. Note that $\mathcal{L}_{prior}$ is typically much smaller than $\mathcal{L}_{ce}$, hence the large $\alpha$.}
\paragraph{+\textsc{SfT}:} Using PyTorch pruning module\footnote{\url{https://pytorch.org/tutorials/intermediate/pruning_tutorial.html}}, we extract the subnetwork from the pretained weights $\theta^0$ following Algorithm \ref{alg:imp} and \textit{Step$_0$} of \textsc{SfT} strategy. More specifically, we consider metrics of the encoder part of the model (see Figure \ref{fig:design-0}), excluding the image and text embeddings as well as the classifier layer in both UC2 and M3P.

Since the network architecture varies between UC2 and M3P, the pruning is applied to a different set of parameters. 
We perform IMP and prune a set of weights with the lowest-magnitude globally throughout the network after each fine-tuning epoch (number of epochs=5). Based on preliminary experiments, we iteratively prune a certain fraction of the lowest-magnitude weights (i.e.\ $p=10\%$) at each epoch which results in the  final sparsity level of around $40\%$ in both models. Considering the exclusion of some parameters, the level of sparsity is 12.28\% for UC2 and 13.44\% for M3P.\footnote{UC2 has around 281.66 M parameters where 85.52 M of them are involved during the pruning process. M3P has 376.90 M parameters and 123.67 M of them are considered for pruning.}
As our focus is not on conducting a large-scale analysis over different sparsity levels, we leave this topic for future work.

Furthermore, following \textit{Step$_1$} of the \textsc{SfT} strategy, we fine-tune the model using the pruning mask $M$.
In both steps, we incorporate similarity-based loss using word embeddings prior (i.e.\ \textit{prior$_{em}$}) in our experiments.
This effectively leads to a  two-stage pruning and sparse fine-tuning process, termed as \textit{With prior$_{em}$+\textsc{SfT}}.
\paragraph{+\textsc{CdM}:}
To perform code-mixing, the English questions and the bilingual dictionaries of MUSE \citep{DBLP:conf/iclr/LampleCRDJ18} are used  as the  basis. We use all 7 target languages in xGQA during the code-mixing augmentation. To perform \textit{With prior$_{em}$+\textsc{SfT}+\textsc{CdM}} experiments, we continue with fine-tuning   by applying the code-mixing during the \textit{Step$_1$} of \textsc{SfT} after pruning the model according to the \textit{Step$_0$} of \textsc{SfT}. We find that including code-mixing during the pruning step (i.e. \textit{Step$_0$}) negatively impacts the model performance in the experiments that follow \textit{With prior$_{em}$+\textsc{SfT}+\textsc{CdM}} strategy.  
\subsection{Baselines and Previous Results:}
We create \textit{Our baselines} by directly evaluating the monolingual fine-tuned models on the test set of the target languages. For each model, we report another baseline using the results in \citet{DBLP:journals/corr/abs-2201-11732}. 
Moreover, we compare our model with a previous study \cite{Liu2022DelvingDI}, where
the low performance of multilingual vision-language models (i.e. UC2 and M3P) in the xGQA dataset has been addressed through sophisticated classification architectures, fine-tuning strategies, and modifications of the model input via question-type conditioning. In addition, we report results of \textit{Translate-Test} setup from \citet{DBLP:journals/corr/abs-2201-11732} where target language test data is translated to English and an English-only fine-tuned model is evaluated on the translated test set.

\section{Results and Discussion}
In this section, we report the results of the different fine-tuning strategies. The proposed strategies result in the best-performing models across all 7 target languages in the cross-lingual visual question answering task. A summary of the results with various strategies is provided in Table \ref{tab:main}.
\paragraph{With prior$_{xx}$:}
Our first set of experimental results shows the advantage of using the proposed loss along with the standard cross-entropy loss for the VQA task.
The proposed strategy (i.e. \textit{With prior$_{xx}$}) improves the average cross-lingual zero-shot transfer accuracy by $+7.99$ and $+8.66$ points over the UC2 baseline using WordNet and GloVe embeddings, respectively. At the same time, it shows gains of $+6.9$ and $+7.6$ absolute accuracy points using different modeling choice (i.e. M3P) with \textit{prior$_{wn}$} and \textit{prior$_{em}$}, respectively.
The results indicate that the similarity-based loss obtained from linguistic priors can effectively guide the models during training. They also support our hypothesis that incorporating additional semantic prior knowledge about the label space improves the cross-lingual generalisation. Among the proposed semantic distances, the GloVe embeddings-based distance delivers the greatest improvements in almost all languages.
One major conceptual difference between our WordNet and GloVe-based distance that could explain this difference in performance is that the former is sparse and heuristic, whereas the latter is dense and continuous. GloVe will also capture relations such as antonym labels (e.g.\ male/female, boy/girl, or yes/no).
\paragraph{With prior$_{em}$+\textsc{SfT}:}
The results demonstrate the importance of a task-specific sparse fine-tuning strategy (i.e. \textsc{SfT}) for adapting the multilingual vision-language models in the downstream VQA task without modifications to the model. The \textsc{SfT} strategy brings further improvements (i.e $+0.55$) over the \textit{With prior$_{em}$} strategy for UC2. Even though it does not surpass the previous strategy in M3P and provides slightly lower performance for some of the target languages in UC2, such as Bangali (Bn) and Indonesian (Id), it yields considerably more stable (lower variance) performance across random seeds in all 7 target languages. 
It is also worth noting that the \textsc{SfT} strategy offers a task-specific and parameter-efficient structure for both models, where a fraction of the encoder's parameters ($12.28$\%
of parameters in UC2 and $13.44$\% of parameters in M3P) are masked and ignored during the fine-tuning. These results suggest that  \textsc{SfT} is successful in discovering language-neutral and task-specific parameters that generalise well cross-lingually for xGQA, similar to the finding by \citet{Foroutan2022DiscoveringLS} for text-only tasks.
\paragraph{With prior$_{em}$+\textsc{SfT}+\textsc{CdM}:}The highest zero-shot transfer performance observed in our experiments is obtained by leveraging the code-mixing strategy on top of the previous best strategy (i.e. With prior$_{em}$+\textsc{SfT}). This strategy achieves much better performance than the previous strategies by a large margin on both transformer models compared to the baselines. The improvement is $+13.12$ and $+12.63$ in average accuracy compared to UC2 and M3P baselines, respectively.
It can be observed that this approach outperforms the previous work by \citet{Liu2022DelvingDI}, across most of the target languages with better performance and lower variance. 
Notably, our final strategy provides $42.19$ versus $39.87$ for UC2 model and $40.00$ versus $36.19$ for M3P model in terms of averaged accuracy across 7 languages. This confirms that our approach can better adapt the multilingual vision-language models for the cross-lingual VQA task.

We further aim to understand the impact of \textsc{CdM} in isolation where we do not perform \textsc{SfT}. 
It can be seen in Table \ref{tab:main}, applying \textsc{CdM} as the only strategy results in a large performance drop for UC2 model in some of the target languages, especially in Indonesian (Id) and Chinese (Zh). It also leads to higher variance compared to its counterpart which only benefits from the \textsc{SfT} strategy in both models.
This result demonstrates synergies between the proposed strategies, with the combination of \textsc{CdM}, which promotes alignment of word representations between source and target languages, and \textsc{SfT}, which discovers subnetworks that may be more language-neutral, achieving a large improvement in combination whereas effects are more moderate (or negative) when applied in isolation.
It is worth to note that we also conduct experiments with only CDM strategy (i.e. excluding the With prior$_{em}$ strategy). However, the results were lower than applying the With prior$_{em}$+CDM (e.g Avg=32.76 compare to Avg=35.16 using UC2).

\addtolength{\tabcolsep}{-3pt}
\begin{table}[tb]
    \centering
    \scalebox{.78}{
    \begin{tabular}{c p{4.5cm}c c c }
    \toprule
     & \textbf{Model} & \multicolumn{3}{c}{\textbf{Avg.}}\\
     \cmidrule(l){3-5}
        &  & w/o Syn. &  w Syn. & Diff.\\
        \midrule
       UC2 & Our Baseline & 29.07 & 
        29.96 & +0.89
       \\
         \cmidrule(lr){2-2} \cmidrule(lr){3-5} 
        & With prior$_{wn}$  & 37.06 &
        38.91 & +1.85\\
          & With prior$_{em}$ & 37.73&
        39.06 & +1.33\\
\cmidrule(lr){2-2} \cmidrule(lr){3-5}
        &  With prior$_{em}$+ \textsc{SfT} & 38.28 &

        39.67 & +1.39\\
          \cmidrule(lr){2-2} \cmidrule(lr){3-5} 
        & With prior$_{em}$ + \textsc{SfT} + \textsc{CdM} &\bf{42.19}&

        \bf{43.90} & +1.71\\
        \midrule
       M3P & Our Baseline & 27.37 &
                
        31.83 & +4.56\\
          \cmidrule(lr){2-2} \cmidrule(lr){3-5} 
        & With prior$_{wn}$ & 34.28&
        
        37.70 & +3.42\\
     & With prior$_{em}$ & 34.97&
        38.85 & +3.88\\
\cmidrule(lr){2-2} \cmidrule(lr){3-5}
     &  With prior$_{em}$ + \textsc{SfT} & 34.02  & 
        38.25& +4.23\\
        \cmidrule(lr){2-2} \cmidrule(lr){3-5} 
        &  With prior$_{em}$ + \textsc{SfT} + \textsc{CdM}   & \bf{40.00}&
                        \bf{43.52}& +3.52  \\
     \bottomrule
    \end{tabular}
   }
    \caption{Results of adjusting the evaluation metric to consider the synonym of
the target label as a correct prediction (\emph{w Syn.}). The \emph{w/o Syn.} column indicates the results before the adjustment.}
    \label{tab:syn}
    \vspace{-0.5cm}
\end{table}
\begin{table*}[tb]
  \resizebox{1\linewidth}{!}{
   \begin{tabular}{lcc|p{.5cm}|c|p{.5cm}|c|p{.5cm}|c|p{.5cm}|c|p{.5cm}|}
 \toprule
     \textbf{Model} & \textbf{Lang.} & \multicolumn{10}{c}{\textbf{5 most-confused labels}}\\
      & & \multicolumn{10}{c}{label:prediction (rel.) } \\
 
        \midrule
       Our Baseline & En & 
  girl:woman (hyp) & 27 &
  material:color (hpo) &23 &
  lady:woman (hyp) &18 &
  coffee table:table (hyp) &17 &
  zebras:zebra (syn) &16 \\

&{Bn} &
  sailboats:sailboat (syn) & 3 &
  skater:skateboarder (hpo) &3 &
  plain:field (syn) &2 &
  trees:tree (syn) &2 &
  tank top:shirt (hyp) &1 \\

&De &
  girl:woman (hyp) & 33 &
  material:color (hpo) &21 &
  lady:woman (hyp) &16 &
  woman:girl (hpo) &13 &
  street sign:sign (hyp) &13 \\

&{Id}	&
  girl:woman (hyp) &28 &
  lady:woman (hyp) &18 &
  skater:skateboarder (hpo) &15 &
  woman:girl (hpo) &14 &
  zebras:zebra (syn) &12 \\

&{Ko}& 
  girl:woman (hyp) &7 &
  skater:skateboarder (hpo) &7 &
  boy:man (hyp) &2 &
  fire truck:truck (hyp) &2 &
  gown:dress (hyp) &1 \\

&{Pt}&
  girl:woman (hyp) &22 &
  skater:skateboarder (hpo) &17 &
  lady:woman (hyp) &13 &
  zebras:zebra (syn) &12 &
  woman:girl (hpo) &11 \\

&{Ru}&
  girl:woman (hyp) & 32 &
  skater:skateboarder (hpo) & 17 &
  lady:woman (hyp) & 17 &
  woman:girl (hpo) & 14 &
  cabinets:cabinet (syn) & 12 \\

& {Zh}&
  girl:woman (hyp) & 26 &
  chairs:chair (syn) & 15 &
  cabinets:cabinet (syn) & 15 &
  skater:skateboarder (hpo) & 15 &
  lady:woman (hyp) & 15
  \\
  \midrule
  Our Best Strategy  & {En} &
  girl:woman (hyp) & 28 &
  material:color (hpo) & 24 &
  cabinets:cabinet (syn) & 20 &
  woman:girl (hpo) & 18 &
  zebras:zebra (syn) & 16 \\
&{Bn} &
  cabinets:cabinet (syn) & 29 &
  girl:woman (hyp) & 19 &
  skater:skateboarder (hpo) & 15 &
  woman:girl (hpo) & 12 &
  lady:woman (hyp) &  12 \\
&De &
  girl:woman (hyp) & 32 &
  material:color (hpo) & 23 &
  lady:woman (hyp) & 18 &
  cabinets:cabinet (syn) & 17 &
  woman:girl (hpo) & 16 \\

&{Id}	&
  girl:woman (hyp) & 27 &
  cabinets:cabinet (syn) & 24 &
  woman:girl (hpo) & 17 &
  chairs:chair (syn) & 17 &
  lady:woman (hyp) & 17 \\

&{Ko}& 
  cabinets:cabinet (syn) & 39 &
  girl:woman (hyp) & 34 &
  elephants:elephant (syn) & 20 &
  woman:girl (hpo) & 17 &
  chairs:chair (syn) & 17 \\

&{Pt}&
  material:color (hpo) & 25 &
  girl:woman (hyp) & 24 &
  woman:girl (hpo) & 20 &
  zebras:zebra (syn) & 15 &
  lady:woman (hyp) & 15 \\

&{Ru}&	
  girl:woman (hyp) & 33 &
  cabinets:cabinet (syn) & 25 &
  material:color (hpo) & 19 &
  woman:girl (hpo) & 18 &
  lady:woman (hyp) & 16 \\
 
& {Zh} & cabinets:cabinet (syn) & 32 & girl:woman (hyp) &27 & chairs:chair (syn) &26 & zebras:zebra (syn) &25 & elephants:elephant (syn) &24 \\
\bottomrule
\end{tabular}%
}
 \caption{ The 5 most-confused labels for each language, specifically where the UC2 model predicts a \emph{synonym} (syn), \emph{hypernym} (hyp), or \emph{hyponym} (hpo) of the target label. The number of “wrong” predictions that are in a synonym/hypernymy/hyponymy relationship (rel.) with the ground truth label is reported in separate columns. 
 }
    \label{tab:syn-details-}
    \vspace{-0.5cm}

\end{table*}

\section{Further Analysis}
To further investigate the effect of synonymy relations among the target labels on xGQA evaluation results,
we modify the evaluation metric to consider synonyms of the ground truth label as a correct prediction. We use the WordNet synonym synset for this purpose. For instance we consider \emph{couch} correct if the ground truth label is \emph{sofa}, or \emph{girls} if the ground truth label is \emph{girl}.
We note that confusion between synonymous labels is relatively common in xGQA; if we consider synonyms to be correct answers, model performance is actually higher than reported by the original accuracy by $0.9$-$1.85$ and $3.42$-$4.56$ percentage points with UC2 and M3P, respectively (see Table \ref{tab:syn}). 

Table \ref{tab:syn-details-} shows the 5 most-confused labels for each language, specifically where the UC2 model predicts a synonym, hypernym, or hyponym of the target label.
While synonyms are predominantly due to inflectional differences (singular/plural), we also find a large number of ``wrong'' predictions that are in a hypernymy/hyponymy relationship with the ground truth, and semantically plausible (\emph{girl} or \emph{lady} vs.\ \emph{woman}).
Although the confusion between similar labels motivated our use of linguistic priors, the performance improvement that we observe is not predominantly due to a reduction in this confusion. In fact, with our best strategy, the number of ``wrong'' predictions that are semantically plausible even increases for UC2, especially for some low-resource languages such as Bengali (Bn) and Korean (Ko), which we take as a positive result: our strict accuracy results in Table \ref{tab:main} already show a substantial improvement for these languages, and with a more permissive evaluation metric, gains over the baseline would be even greater.
Similar results are observed when we only take into account synonymy relationships. 

\section{Related Works}
The primary motivation for this work is the low cross-lingual generalization of multilingual vision-language pretrained models. There are a number of works addressing this problem. \citet{Zeng2022CrossViewLM} introduce cross-view language modeling by considering both image-caption pairs and parallel sentence pairs as two different views of the same object and train the model to align the two views by maximizing the mutual information between them with conditional masked language modeling and contrastive loss. Whereas they report a state-of-the-art zero-shot cross-lingual performance for xGQA, their method 
demands a pretraining step as well as high computing resources and multilingual language-vision datasets. In contrast, our proposed strategy can be applied on top of any multilingual vision-language pretrained model as an adaptation step.
Our approach is similar to the work by \citet{Liu2022DelvingDI}, where they propose a set of methods that improves previously low transfer performance and thus substantially reduce the gap to monolingual English performance.  However, their approach is more complex and our final strategy provides better performance with a sparse encoder.\\
\textbf{Similarity-based loss:}
There is an increasing interest in incorporating prior domain knowledge in neural NLP downstream tasks. Prior knowledge of the language has been applied recently to language generation learning. \citet{Li2020DatadependentGP} introduces a technique that imposes the prior from (linguistic) data over the sequence prediction models and improves performance in typical language generation tasks, including machine translation, text summarization, and image captioning. \citet{chousa2018training} propose a novel NMT loss function that includes word similarity in forms of distances in a word embedding space and it leads to a substantial gain in the machine translation task.\\
\textbf{Sparse fine-tuning:}
Our approach is inspired by studies of sparse fine-tuning methods \citep{ansell-etal-2022-composable,Liang2021FindingSS, Foroutan2022DiscoveringLS} .
\citet{ansell-etal-2022-composable} and \citet{Liang2021FindingSS} claim that non-restricted fine-tuning of multilingual models is prone to over-fitting on source language as well as catastrophic forgetting. They suppose that parameter interference is one of the causes of this degradation. \citet{Foroutan2022DiscoveringLS} suggest that language-specific and language-neutral subnetworks play a prominent role in the cross-lingual generalisation of the multilingual language model (i.e Multilingual BERT).
In this work, we follow the above-mentioned ideas by looking at the structure and weights of multilingual vision-language models in the VQA task.\\
\textbf{Code-Switching:}
Data augmentation training using Code-Switching offers a significant improvement to the low-resource languages. It helps the model explicitly learn the relationship among words in different languages.
It has been applied to the training of various multimodal multilingual  models such as M3P \cite{9577347}  and CCLM \cite{Zeng2022CrossViewLM}.
\citet{raj-khan-etal-2021-towards-developing} create a  multilingual and code-mixed VQA dataset in eleven different language setups considering the multiple Indian and European languages as well as their code-mixed  versions. They propose a knowledge distillation  to  extend an English language-vision  model (teacher) into a multilingual and code-mixed  model (student). However, this dataset is not diverse as xGQA in terms of covering the low-resource languages.
\section{Conclusion}
We present a series of strategies to fine-tune multilingual vision-language pretrained models for better cross-lingual generalisation in the visual question answering task. 
Our approach is based on various adaptation techniques aimed to mitigate the number of issues that we discovered regarding the training and evaluation of multilingual vision-language models on xGQA. 
Comparing our approach with the baseline and previous similar work in several pretrained models, the results indicate substantial improvements across target languages. 
The improvement is $+13.12$ and $+12.63$ in average accuracy over all 7 languages in xGQA compared to UC2 and M3P baselines, respectively.

We perform an analysis of closely related target labels in xGQA, proposing a new metric that rewards synonymous predictions and further demonstrates the success of the proposed strategies.
This analysis also highlights the need for future research on the label space and evaluation metrics for cross-lingual VQA.

\section*{Acknowledgments}
We would like to thank Xin Sennrich and  Alham Fikri Aji for their helpful feedback on language resources. This work was funded by the Swiss National Science Foundation (project MUTAMUR; no. 176727) at the University of Zurich.

\bibliography{aaai23}

\end{document}